\documentclass[conference]{IEEEtran}
\IEEEoverridecommandlockouts
\usepackage{cite}
\usepackage{amsmath,amssymb,amsfonts}
\usepackage{multirow}
\usepackage{algorithmic}
\usepackage{graphicx}
\usepackage{textcomp}
\usepackage{xcolor}

\def\BibTeX{{\rm B\kern-.05em{\sc i\kern-.025em b}\kern-.08em
    T\kern-.1667em\lower.7ex\hbox{E}\kern-.125emX}}
\begin{document}

\title{Multi-modal Imputation for \\ Alzheimer's Disease Classification}

\author{\IEEEauthorblockN{Abhijith Shaji}
\IEEEauthorblockA{\textit{Information Sciences Institute} \\
\textit{University of Southern California}\\
Los Angeles, CA, USA \\
\texttt{ashaji@isi.edu}}
\and
\IEEEauthorblockN{Tamoghna Chattopadhyay}
\IEEEauthorblockA{\textit{University of Southern California} \\
Los Angeles, CA, USA \\
\texttt{tchattop@usc.edu}}
\and
\IEEEauthorblockN{Sophia I. Thomopoulos}
\IEEEauthorblockA{\textit{University of Southern California} \\
Los Angeles, CA, USA \\
\texttt{sthomopo@usc.edu}}
\and
\IEEEauthorblockN{Greg Ver Steeg}
\IEEEauthorblockA{\textit{University of California} \\
Riverside, CA, USA \\
\texttt{greg.versteeg@ucr.edu}}
\and
\IEEEauthorblockN{Paul M. Thompson}
\IEEEauthorblockA{\textit{Stevens Neuroimaging and Informatics Institute} \\
\textit{University of Southern California}\\
Los Angeles, CA, USA \\
\texttt{pthomp@usc.edu}}
\and
\IEEEauthorblockN{Jose-Luis Ambite}
\IEEEauthorblockA{\textit{Information Sciences Institute} \\
\textit{University of Southern California}\\
Los Angeles, CA, USA \\
\texttt{ambite@isi.edu}}
}

\maketitle

\begin{abstract}
Deep learning has been successful in predicting neurodegenerative disorders, such as Alzheimer's disease, from magnetic resonance imaging (MRI). Combining multiple imaging modalities, such as T1-weighted (T1) and diffusion-weighted imaging (DWI) scans, can increase diagnostic performance. However, complete multimodal datasets are not always available. 
We use a conditional denoising diffusion probabilistic model to impute missing DWI scans from T1 scans. 
We perform extensive experiments to evaluate whether such imputation improves the accuracy of uni-modal and bi-modal deep learning models for 3-way Alzheimer’s disease classification (cognitively normal, mild cognitive impairment, and Alzheimer’s disease). We observe improvements in several metrics, particularly those sensitive to minority classes, for several imputation configurations. 
\end{abstract}


\section{Introduction}
\label{sec:intro}
Neurodegenerative disorders such as Alzheimer's disease (AD) is characterized by progressive anatomical and microstructural changes in the brain that can be tracked using modern neuroimaging techniques. In particular, multiple magnetic resonance imaging (MRI) modalities are often combined to provide complementary information; T1-weighted MRI characterizes the structural anatomy of the brain, whereas diffusion-weighted imaging (DWI) provides information on microstructural changes through diffusion measures such as fractional anisotropy and mean diffusivity \cite{basser1994mr, jones2008studying, mori2006principles, feng2025microstructural}. Consequently, multimodal learning frameworks that integrate these contrasts often outperform their unimodal counterparts in disease classification and prognostic tasks. 

However, complete multimodal datasets are not always available. DWI acquisitions are typically longer and prone to motion and susceptibility artifacts \cite{topolnjak2024assessment}. This can lead to incomplete datasets where T1-weighted scans are widely available, but the DWI scans are missing for a large subset of the cohort. This incompleteness reduces the effective size of the multimodal dataset, thus impeding generalization and model stability for the multimodal networks. Traditional imputation approaches fail to capture the complex, non-linear relationship between complementary modalities, motivating the use of generative approaches.
Recent advances in image generation architectures, including denoising diffusion probabilistic models (DDPM), latent diffusion models, and flow models, can learn highly expressive generative priors for medical image synthesis and cross-modality translation \cite{ho2020denoising,dorjsembe2022three,pinaya2022brain,zhang2025diffusion}. 

In this paper, we use a conditional 3D DDPM as an image translator to impute missing DWI scans from corresponding 3D T1-weighted MRIs, and generate larger multimodal datasets.
We evaluate the impact of this imputation on the downstream task of Alzheimer’s disease 3-way classification between cognitively normal individuals, those having mild cognitive impairment, and Alzheimer’s disease.

\section{Imputation using Image Translation}
\label{sec:imputation}
Multimodal neural networks integrate complementary information from different data sources, providing richer and more robust feature representations compared to the unimodal setup \cite{baltruvsaitis2018multimodal}. In neuroimaging, combining T1-weighted structural MRIs with diffusion-weighted images provides both anatomical and microstructural assessments of the brain, potentially improving diagnostic and predictive performance.

However, this increased representational capacity comes at the cost of higher model complexity. Multimodal architectures typically have more trainable parameters, arising from the additional modality-specific encoders and fusion layers, which substantially increase the amount of data required to effectively train the model \cite{ramachandram2017deep}.

When the training data is limited or incomplete, the models could underperform and may not be able to exploit the theoretical advantages of multimodal fusion, sometimes even performing worse than their unimodal counterparts.
Many large datasets often suffer from incomplete or missing data points in a multimodal dataset. These incomplete datapoints lead to a substantial loss of usable data, forcing analyses to rely on unimodal subsets or a smaller multimodal subset with all the required modalities available.

To mitigate this limitation, we propose to impute missing imaging modalities using a conditional generative framework based on denoising diffusion probabilistic models (DDPMs). The model is trained to learn the conditional distribution $\Pr(\text{DWI} \mid \text{T1})$ from paired scans and can be further used to generate anatomically consistent synthetic DWI volumes for subjects with only T1 data available. Incorporating these imputed volumes allows us to use the previously unusable datapoints in multimodal analyses.


We evaluated the impact of the augmented multimodal dataset on the downstream task of Alzheimer’s disease classification across three diagnostic categories: cognitively normal (CN), mild cognitive impairment (MCI), and Alzheimer’s disease (AD). Incorporating the imputed DWI data enabled the inclusion of more scans and thus effectively increased the size of the multimodal dataset. 
We compared the performance of models trained with additional DDPM-generated images against three baselines: a model trained only on the limited paired subset, a blank-imputation baseline, and a diagnosis-average imputation baseline. In the blank imputation baseline, the missing DWI volumes are replaced with zero-valued images, providing a lower-bound reference that contains no diffusion information. In the diagnosis-average imputation baseline, each missing DWI scan is replaced with the mean DWI volume computed across subjects of the same diagnosis group (CN, MCI, or AD).
We show a comprehensive evaluation of uni-modal and bi-modal models for AD classification with different amounts and distribution of imputed DWI volumes.

\section{Data and Preprocessing}
\label{sec:data}

We used MRI scans from the first three phases of Alzheimer’s Disease Neuroimaging Initiative (ADNI) \cite{jack2008alzheimer}. We focused on two available sub-modalities of MRI: T1-weighted (T1) and the fractional anisotropy (FA) scalar mapping of diffusion-weighted imaging (DWI).

Diffusion-weighting is a technique employed by Diffusion Tensor Imaging (DTI) to model brain microstructure \textit{in vivo}. The diffusion tensor model approximates local diffusion using a spatially varying tensor, represented by a 3D Gaussian at each voxel. DTI is summarized using four scalar maps: fractional anisotropy (FA), axial, mean, and radial diffusivity (AxD, MD, RD). These mappings are used to characterize the shape of the diffusion tensor at each voxel, derived from its three principal eigenvalues indicating the primary directions of water diffusion at each voxel. We focus on imputing FA scalar maps (\ref{eq}):

\begin{equation}
\mathrm{FA}
=
\sqrt{\frac{1}{2}}
\;
\frac{
\sqrt{
(\lambda_1 - {\lambda_2})^2 +
(\lambda_2 - {\lambda_3})^2 +
(\lambda_3 - {\lambda_1})^2
}
}{
\sqrt{
\lambda_1^2 + \lambda_2^2 + \lambda_3^2
}
}
\label{eq}
\end{equation}
where $\lambda_1$, $\lambda_2$ and $\lambda_3$ are the principal eigenvalues.

T1-weighted MRI volumes were preprocessed using N4 bias correction, skull stripping, nonlinear registration to an in-house template of 9 degrees of freedom (DOF), and resampling to 2 mm isotropic resolution (final size = $91 \times 109 \times 91$ voxels). Images were min–max scaled to [0, 1]. DWI MRI from ADNIGO/2 followed the preprocessing in \cite{zavaliangos2019diffusion} \cite{nir2013effectiveness} , while ADNI3 used an updated pipeline accounting for scanner and protocol variability \cite{thomopoulos2021diffusion}. Diffusion tensor imaging (DTI) scalar maps were computed using FSL’s DTIFIT \cite{woolrich2009bayesian} \cite{jenkinson2012fsl}, resampled to 2 mm isotropic ($91 \times 109 \times 91$ voxels), and aligned with the ENIGMA TBSS DTI template \cite{jahanshad2013multi} \cite{chattopadhyay2023predicting}.
Specifically, the T1-weighted images correspond to accelerated T1 scans that were registered to a common template using a 9-DOF spatial transform, while the DWI-derived scalar FA maps were computed from the DTI data. Every subject with an available DWI scan also had a corresponding T1 scan.


Based on availability, we defined three datasets: unimodal T1s, unimodal DWIs, and bimodal T1+DWIs. 

The \textbf{T1-only dataset} was split into training, validation, and test sets containing 3901, 841, and 859 scans, respectively (average age $73.92\pm7.49$; $2700$ females, $2901$ males).
The distribution of diagnostic categories was Training Set: CN = 1505, MCI = 1821, AD = 575; Validation Set: CN = 317, MCI = 403, AD = 121; and Test Set: CN = 294, MCI = 398, AD = 167. 

The \textbf{T1+DWI dataset} was split into training, validation, and test sets containing 642, 137, and 137 scans, respectively (average age $74.67\pm7.81$; $468$ females, $448$ males).
The distribution of diagnostic categories was Training Set: CN = 372, MCI = 206, AD = 64; Validation Set: CN = 80, MCI = 44, AD = 13; and Test Set: CN = 80, MCI = 44, AD = 13.
The T1+DWI splits were proper subsets of the corresponding T1-only splits. 

The \textbf{DWI-only dataset} was just the DWI portion of the T1+DWI datasets with the same splits. 

Although the datasets contained repeated scans from some subjects, the splits were defined to be subject-disjoint for all training, validation, and test sets. 

\section{Deep Learning Architectures}

\subsection{3D DDPM Architecture}
\begin{figure}[htbp]
\centering
\includegraphics[width=\columnwidth]{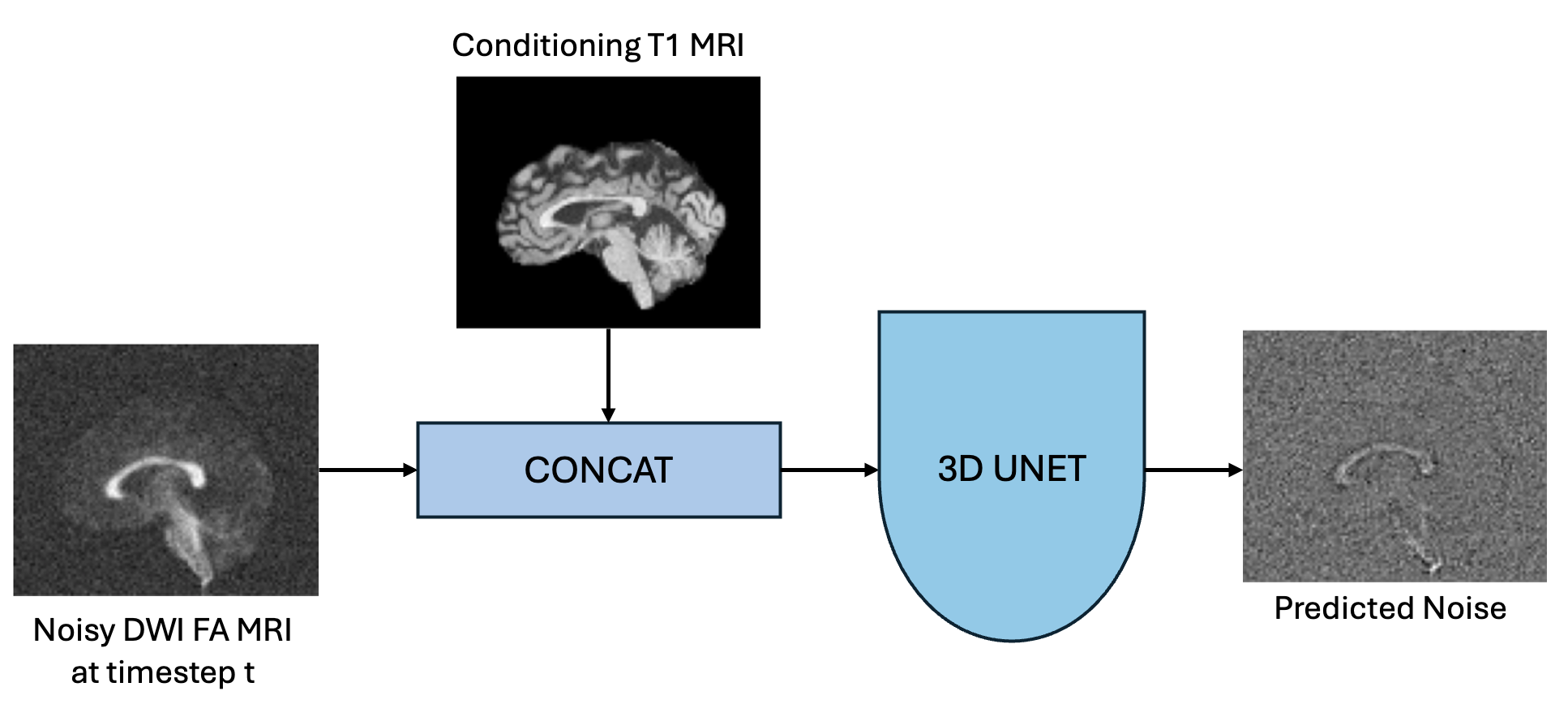}
\caption{3D UNet architecture used to predict the noise output from a noisy DWI FA MRI conditioned on T1 MRI.}
\label{fig:DDPM}
\end{figure}

We built upon the 3D U-Net based diffusion model architecture from the MONAI-generative models \cite{cardoso2022monai,pinaya2023generative}. The network was designed to operate on volumetric MRI data, taking as input two channels corresponding to T1-weighted and diffusion weighted images, and producing a single channel output of the predicted noise component at a given diffusion timestep as illustrated in Fig.~\ref{fig:DDPM}. 

The model consists of an encoder-decoder U-Net architecture \cite{ronneberger2015u} with residual connections, temporal conditioning, and attention mechanisms for volumetric feature extraction.
The encoder comprises three hierarchical downsampling stages with feature widths of 128, 128, and 256 channels, respectively. The final encoder stage incorporates self-attention layers to capture long-range spatial dependencies across the 3D feature space.
The decoder mirrors the encoder, employing three upsampling stages with learned 3D convolutions and skip connections that concatenate corresponding encoder features. 
This configuration results in a symmetric 3D residual U-Net architecture with attention mechanism, optimized for denoising within a diffusion probabilistic modeling framework for volumetric MRI data synthesis.

To incorporate anatomical information, the denoising network conditions on the T1 scan by concatenating it with the noisy DWI input along the channel dimension. Rather than predicting noise for DWI alone, the model learns to predict noise for the combined T1–DWI input, helping it generate DWI volumes that remain consistent with the subject’s underlying T1 anatomy.
The 3D DDPM trained on the paired dataset (642 train + 137 validation) 
when evaluated on the 137 test images, achieved a mean SSIM-3D of 0.36, PSNR of 23.61, L1 error of 0.043, and MSE of 0.0057. 

\begin{figure}[htbp]
\centering
\includegraphics[width=\columnwidth]{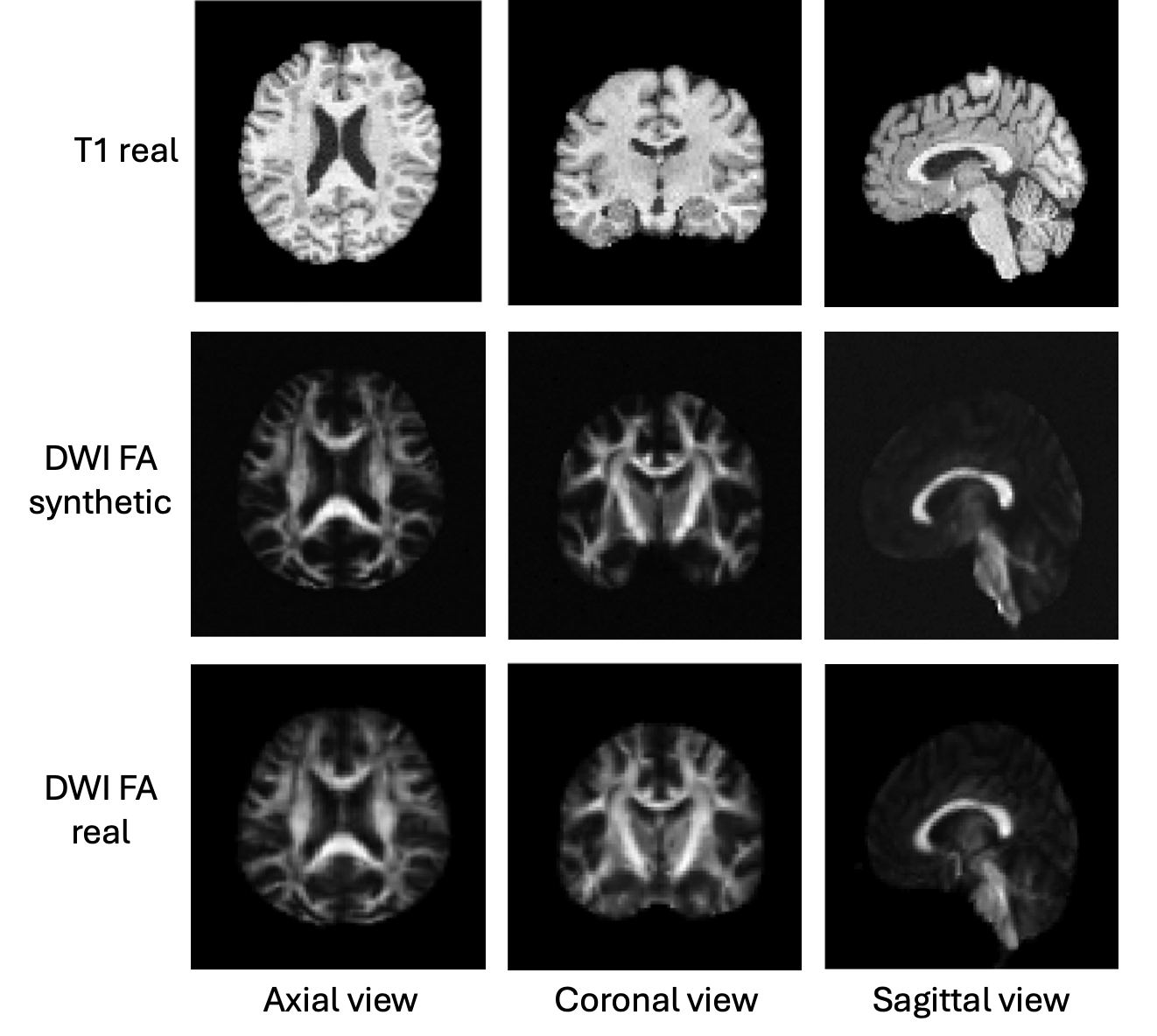}
\caption{T1-weighted MRI, synthetic DWI FA MRI, and real DWI FA MRI across axial, coronal, and sagittal views for a single healthy subject.}
\label{fig:mri}
\end{figure}

Fig.~\ref{fig:mri} shows an example of a T1 scan, the corresponding paired DWI FA MRI, and the conditionally generated DWI FA MRI. 

\subsection{Uni-Modal 3D CNN Architecture}
\label{sec:3dcnn_architecture}

\begin{figure}[htbp]
\centering
\includegraphics[width=\columnwidth]{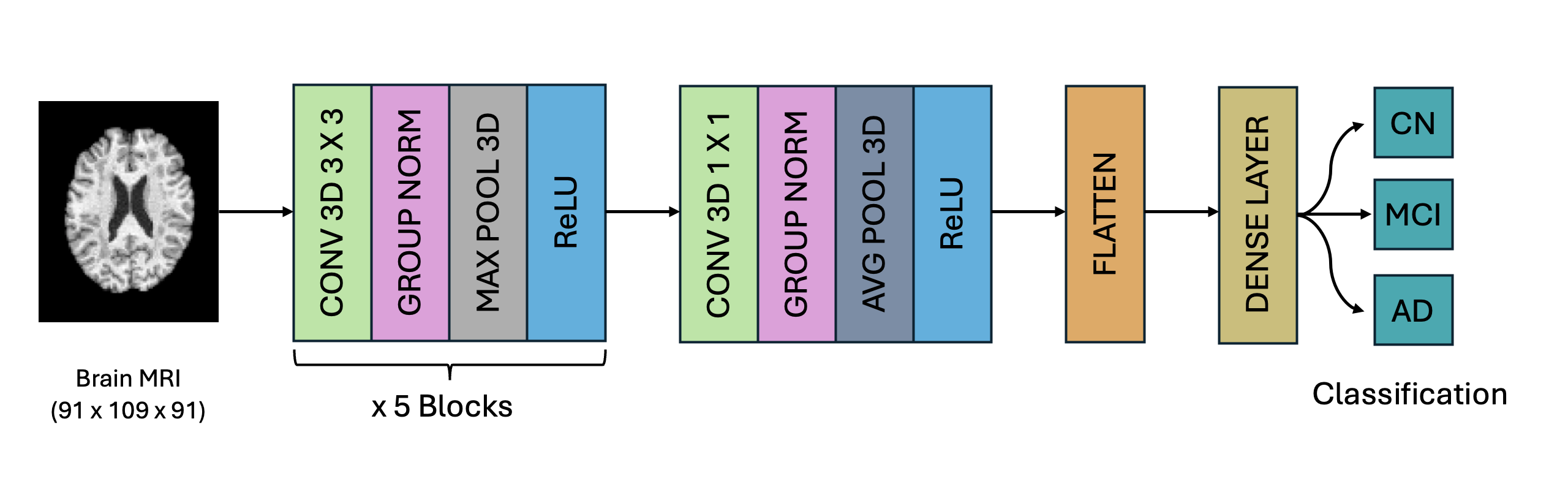}
\caption{Unimodal 3D CNN Architecture.}
\label{fig:unimodal}
\end{figure}

We implemented a 3D convolutional neural network (CNN) to classify Alzheimer’s disease using MRI as the input (Fig.~\ref{fig:unimodal}). The network consisted of 5 convolutional blocks, each comprising a 3D convolutional layer, followed by group normalization layer, 3D max pooling layer and a relu activation layer. The number of filters increased progressively across layers ($32 \rightarrow 64 \rightarrow 128 \rightarrow 256 \rightarrow 256$).
A final $1 \times 1 \times 1$ convolution with 64 filters, followed by group normalization, ReLU activation and average pooling was applied. A dropout layer with a dropout rate of 0.2 was used to prevent overfitting. The resulting feature map was flattened and passed as an input into a softmax classifier. The same model architecture was used for both the T1-only and DWI-only classification experiments.

\subsection{Bi-Modal 3D CNN Architecture} 
\begin{figure}[htbp]
\centering
\includegraphics[width=\columnwidth]{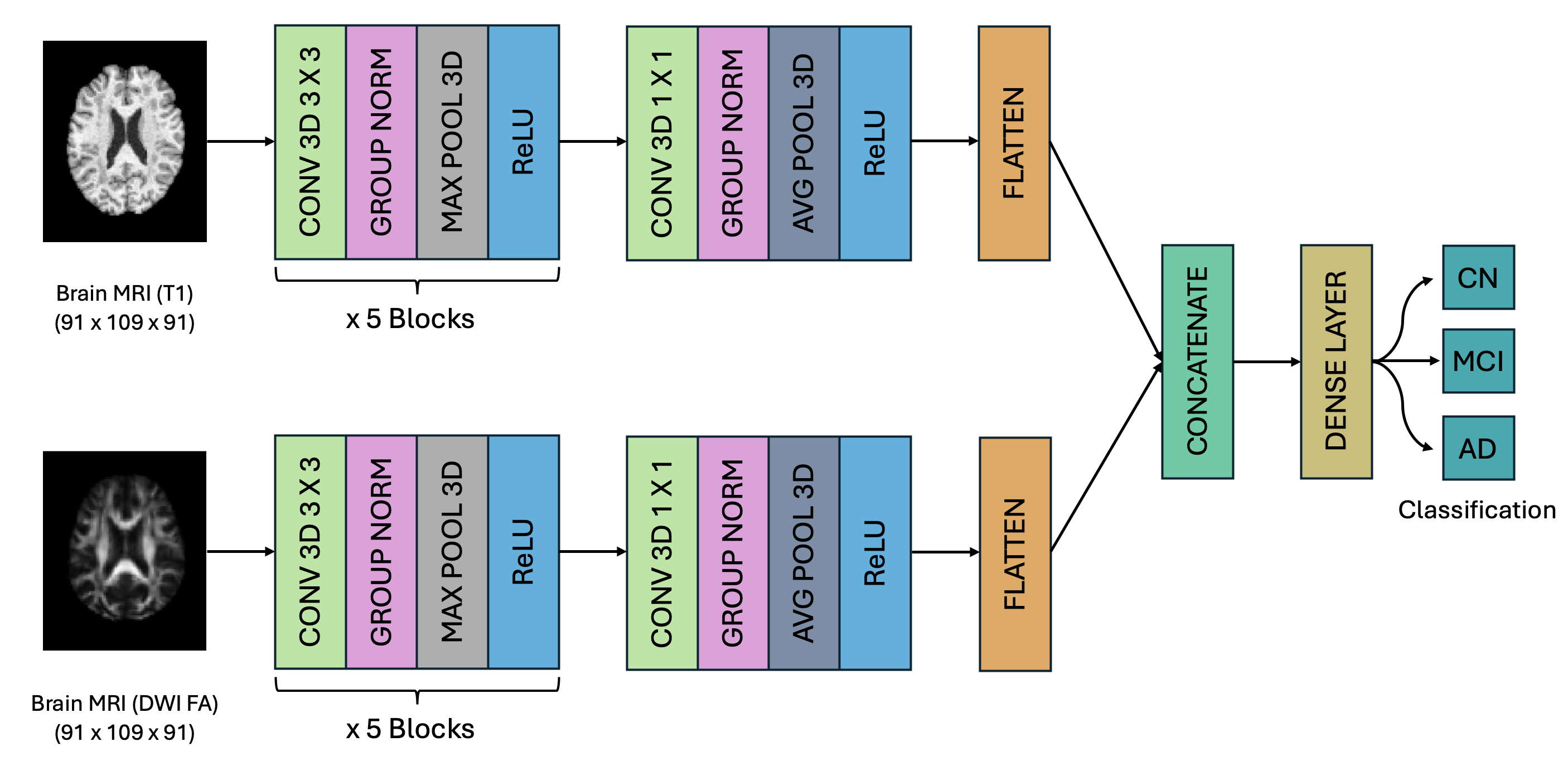}
\caption{Bimodal 3D CNN Architecture.}
\label{fig:bimodal}
\end{figure}

The bimodal 3D CNN is built as an extension to the unimodal 3D CNN described in Section~\ref{sec:3dcnn_architecture} to jointly model T1-weighted and diffusion weighted MRI data (Fig.~\ref{fig:bimodal}). The network comprised of two parallel branches, each identical in structure to the unimodal network, independently extracting features from each modality. The obtained output features from both the branches were then flattened and concatenated in a late fusion manner to form a joint multimodal representation. The fused feature vector was then passed as an input into a softmax classifier to predict the diagnostic category.

\subsection{Training the models}

\paragraph{3D CNNs} Both the unimodal and bimodal classifiers used Bayesian Optimization to obtain the optimal hyperparameters for learning rate and weight decay. The search space included learning rates $\{1\times10^{-4}, 1\times10^{-5}, 5\times10^{-5}, 1\times10^{-6}\}$ and weight decay values $\{1\times10^{-4}, 1\times10^{-5}, 1\times10^{-6}\}$. The optimization process selected the best-performing configuration based on validation accuracy \cite{snoek2012practical}.

We trained the model for 100 epochs with early stopping (patience = 10) to prevent overfitting \cite{prechelt2002early}, using the AdamW optimizer \cite{loshchilov2017decoupled} with the optimal hyperparameter configuration, and categorical cross-entropy loss function to optimize multi-class classification performance.


\paragraph{Conditional DDPM}
The DDPM was trained to generate DWI volumes conditioned on T1-weighted MRI using the noise prediction formulation of diffusion models \cite{ho2020denoising}. The diffusion process employed 1,000 timesteps with a scaled linear beta schedule \cite{ho2020denoising,nichol2021improved}, where the noise variance increased from $\beta_{\text{start}}=5\times10^{-4}$ to $\beta_{\text{end}}=1.95\times10^{-2}$. Conditioning was implemented via channel-wise concatenation of the T1 image with the noisy DWI input.
The model was trained for 300 epochs using the Adam optimizer \cite{kingma2014adam} with a learning rate of $5\times10^{-5}$ on the mean squared error loss between the predicted and true noise. Mixed-precision training was used to improve computational efficiency \cite{micikevicius2017mixed}.

The DDPM was trained and sampled using the Quadro RTX 8000 GPUs and the 3D CNNs were trained and evaluated using the NVIDIA GeForce GTX 1080 Ti GPUs.

\section{Results}

Table~\ref{tab:results} summarizes the classification performance across multiple conditional imputation settings using diffusion models (DDPMs), along with baseline strategies of blank, and average diagnosis-based (AvgDX) imputations. We performed the baseline strategies of imputation only on the top four models in the bimodal (T1+DWI) and DWI-only settings.

\begin{table*}[htbp]
\caption{Performance of conditional imputation experiments on Alzheimer's Disease classification.}
The \textit{Imputation} column indicates the strategy used for missing DWI data: 
\textbf{DDPM} denotes diffusion-based imputation, 
\textbf{Blank} replaces missing DWI with zero-valued volumes, 
\textbf{AvgDX} replaces missing DWI with the diagnosis-wise mean DWI volume, and 
\textbf{--} indicates no imputation (only real data used).
The \textit{CN}, \textit{MCI}, and \textit{AD} columns denote the number of training samples added per diagnostic class. \textit{Total} column reports the size of the training set: real MRIs + imputed MRIs.\\

\centering
\small
\resizebox{\textwidth}{!}{%
\begin{tabular}{|c|c|c|c|c|c|c|c|c|c|c|c|}
\hline
\textbf{Experiment} &
\textbf{CN} &
\textbf{MCI} &
\textbf{AD} &
\textbf{Total} &
\textbf{Imputation} &
\textbf{Acc} &
\textbf{Bal Acc} &
\textbf{Micro AUC} &
\textbf{Macro AUC} &
\textbf{Macro Prec} &
\textbf{Macro F1} \\
\hline

\hline
\multirow{25}{*}{%
\parbox{3.2cm}{\centering
\textbf{T1 + DWI}\\
\textbf{(Train=642, Val=137, Test=137)}
}} 
& 0 & 0 & 0 & 642 & -- & 68.03$\pm$2.33 & 46.13$\pm$2.26 & 84.99$\pm$1.43 & 80.39$\pm$2.49 & 42.30$\pm$1.88 & 43.92$\pm$2.20 \\
\cline{2-12}
& 0 & 100 & 50 & 792 & DDPM & 67.88$\pm$2.35 & 47.01$\pm$3.30 & 86.01$\pm$1.15 & 82.16$\pm$1.87 & 55.13$\pm$17.05 & 46.29$\pm$4.95 \\
\cline{2-12}
& 0 & 100 & 100 & 842 & DDPM & 65.11$\pm$1.49 & 47.25$\pm$5.02 & 85.64$\pm$0.50 & 82.14$\pm$0.81 & 52.42$\pm$12.33 & 46.99$\pm$6.75 \\
\cline{2-12}
& 100 & 100 & 100 & 942 & DDPM & 67.15$\pm$1.96 & 50.48$\pm$5.72 & 85.60$\pm$1.29 & 81.06$\pm$2.20 & 60.84$\pm$10.52 & 50.84$\pm$5.97 \\
\cline{2-12}
& 0 & 200 & 100 & 942 & DDPM & \textbf{70.36$\pm$2.38} & 54.52$\pm$6.85 & \textbf{87.04$\pm$1.92} & 83.18$\pm$3.03 & 68.56$\pm$8.08 & 55.75$\pm$7.40 \\
\cline{2-12}
& 0 & 200 & 200 & 1042 & DDPM & 67.74$\pm$3.34 & 52.10$\pm$8.37 & 85.96$\pm$1.99 & 81.70$\pm$3.53 & 65.10$\pm$13.73 & 52.75$\pm$10.47 \\
\cline{2-12}
& 0 & 166 & 308 & 1116 & DDPM & 66.86$\pm$3.47 & 55.86$\pm$7.54 & 85.71$\pm$1.07 & 81.11$\pm$2.32 & 61.92$\pm$10.94 & 54.90$\pm$8.75 \\
\cline{2-12}
& 0 & 250 & 250 & 1142 & DDPM & 67.01$\pm$3.47 & 51.44$\pm$8.96 & 85.93$\pm$1.16 & 82.35$\pm$1.40 & \textbf{70.44$\pm$3.59} & 52.61$\pm$11.29 \\
\cline{2-12}
& 200 & 200 & 200 & 1242 & DDPM & 68.61$\pm$2.01 & 58.14$\pm$6.65 & 86.75$\pm$0.41 & 82.55$\pm$1.48 & 67.62$\pm$4.73 & 57.64$\pm$7.34 \\
\cline{2-12}
& 250 & 250 & 250 & 1392 & DDPM & 67.30$\pm$4.36 & 55.33$\pm$10.27 & 86.36$\pm$1.03 & 82.83$\pm$1.20 & 57.67$\pm$13.47 & 54.38$\pm$13.11 \\
\cline{2-12}
& 100 & 266 & 408 & 1416 & DDPM & 68.03$\pm$0.97 & 55.50$\pm$2.43 & 86.77$\pm$0.34 & 82.95$\pm$0.72 & 70.41$\pm$4.94 & 57.81$\pm$1.62 \\
\cline{2-12}
& 300 & 300 & 300 & 1542 & DDPM & 65.84$\pm$2.14 & 54.55$\pm$2.59 & 86.73$\pm$0.68 & \textbf{83.80$\pm$1.10} & 65.64$\pm$9.97 & 53.95$\pm$3.37 \\
\cline{2-12}
& 350 & 350 & 350 & 1692 & DDPM & 68.32$\pm$2.98 & \textbf{59.96$\pm$4.41} & 85.76$\pm$3.04 & 80.64$\pm$5.04 & 60.60$\pm$11.99 & \textbf{58.67$\pm$7.55} \\
\cline{2-12}
& 200 & 366 & 508 & 1716 & DDPM & 66.72$\pm$1.50 & 55.10$\pm$4.61 & 86.51$\pm$0.74 & 83.22$\pm$1.45 & 64.36$\pm$5.79 & 54.00$\pm$4.28 \\
\cline{2-12}
& 400 & 400 & 400 & 2242 & DDPM & 67.88$\pm$3.45 & 57.20$\pm$3.83 & 86.75$\pm$0.82 & 82.90$\pm$1.42 & 62.72$\pm$5.50 & 57.97$\pm$5.35 \\
\cline{2-12}
& 0 & 200 & 100 & 942 & Blank & 67.01$\pm$0.85 & 50.19$\pm$5.88 & 86.60$\pm$1.33 & 83.17$\pm$2.06 & 59.43$\pm$15.85 & 51.06$\pm$8.32 \\
\cline{2-12}
& 0 & 350 & 350 & 1342 & Blank & 65.11$\pm$2.50 & 52.01$\pm$5.24 & 85.47$\pm$0.69 & 82.08$\pm$0.43 & 63.09$\pm$9.45 & 51.18$\pm$6.40 \\
\cline{2-12}
& 100 & 266 & 408 & 1416 & Blank & 67.45$\pm$1.19 & 56.89$\pm$6.21 & 86.41$\pm$0.61 & 82.52$\pm$1.13 & 69.30$\pm$3.90 & 59.15$\pm$5.29 \\
\cline{2-12}
& 350 & 350 & 350 & 1692 & Blank & 68.03$\pm$3.53 & 55.88$\pm$8.96 & 85.52$\pm$1.71 & 80.98$\pm$3.73 & 56.77$\pm$13.13 & 54.22$\pm$10.86 \\
\cline{2-12}
& 0 & 200 & 100 & 942 & AvgDX & 66.13$\pm$1.19 & 50.06$\pm$4.84 & 85.32$\pm$0.67 & 81.02$\pm$1.03 & 57.76$\pm$10.28 & 48.50$\pm$2.53 \\
\cline{2-12}
& 0 & 350 & 350 & 1342 & AvgDX & 67.01$\pm$2.55 & 48.52$\pm$3.24 & 85.55$\pm$1.33 & 81.31$\pm$1.67 & 58.83$\pm$11.52 & 48.73$\pm$3.72 \\
\cline{2-12}
& 100 & 266 & 408 & 1416 & AvgDX & 65.40$\pm$4.25 & 49.46$\pm$9.37 & 83.67$\pm$2.15 & 79.33$\pm$2.84 & 48.91$\pm$12.76 & 48.06$\pm$11.69 \\
\cline{2-12}
& 350 & 350 & 350 & 1692 & AvgDX & 65.11$\pm$3.59 & 44.30$\pm$5.67 & 84.06$\pm$0.98 & 80.70$\pm$1.86 & 43.11$\pm$7.27 & 42.18$\pm$6.80 \\
\hline

\multirow{25}{*}{%
\parbox{3.2cm}{\centering
\textbf{DWI Only}\\
\textbf{(Train=642, Val=137, Test=137)}
}} 
& 0   & 0   & 0   & 642 & --    & 62.19$\pm$1.63 & 42.28$\pm$3.61 & 83.25$\pm$1.12 & 78.90$\pm$1.54 & 46.28$\pm$6.68 & 41.00$\pm$5.20 \\
\cline{2-12}
& 0   & 100 & 50  & 792 & DDPM  & 63.50$\pm$1.90 & 43.93$\pm$2.74 & 84.38$\pm$0.78 & 79.20$\pm$1.03 & 42.59$\pm$7.29 & 42.49$\pm$3.85 \\
\cline{2-12}
& 0   & 100 & 100 & 842 & DDPM  & 62.63$\pm$2.03 & 41.86$\pm$2.92 & 84.00$\pm$0.62 & 79.50$\pm$0.75 & 42.93$\pm$8.14 & 40.12$\pm$4.38 \\
\cline{2-12}
& 100 & 100 & 100 & 942 & DDPM  & 62.48$\pm$2.15 & 44.71$\pm$2.53 & 82.87$\pm$0.79 & 77.74$\pm$1.65 & 47.08$\pm$5.07 & 44.41$\pm$3.23 \\
\cline{2-12}
& 0   & 200 & 100 & 942 & DDPM  & 63.36$\pm$2.33 & 41.14$\pm$2.87 & 83.98$\pm$1.33 & 79.09$\pm$1.74 & 38.57$\pm$2.02 & 38.76$\pm$3.28 \\
\cline{2-12}
& 0   & 200 & 200 & 1042 & DDPM  & 63.94$\pm$1.88 & 42.91$\pm$2.44 & \textbf{84.54$\pm$1.29} & \textbf{80.11$\pm$1.92} & 39.34$\pm$2.01 & 40.73$\pm$2.38 \\
\cline{2-12}
& 0   & 166 & 308 & 1116 & DDPM  & 62.19$\pm$1.63 & 42.22$\pm$2.36 & 83.36$\pm$0.77 & 78.72$\pm$1.03 & 46.63$\pm$13.99 & 40.82$\pm$3.14 \\
\cline{2-12}
& 0   & 250 & 250 & 1142 & DDPM  & \textbf{65.40$\pm$2.19} & \textbf{51.68$\pm$7.06} & 83.59$\pm$1.43 & 79.18$\pm$1.93 & \textbf{57.18$\pm$9.07} & \textbf{50.79$\pm$5.89} \\
\cline{2-12}
& 0   & 300 & 300 & 1242 & DDPM  & 60.44$\pm$1.69 & 38.70$\pm$2.80 & 83.52$\pm$0.81 & 78.88$\pm$1.34 & 38.03$\pm$4.96 & 36.55$\pm$3.78 \\
\cline{2-12}
& 250 & 250 & 250 & 1392 & DDPM  & 64.38$\pm$1.98 & 43.27$\pm$2.38 & 84.12$\pm$1.21 & 79.74$\pm$1.40 & 52.48$\pm$16.39 & 42.07$\pm$3.17 \\
\cline{2-12}
& 100 & 266 & 408 & 1416 & DDPM  & 63.80$\pm$2.64 & 45.96$\pm$5.13 & 83.75$\pm$1.37 & 79.55$\pm$1.54 & 48.06$\pm$9.61 & 45.34$\pm$6.83 \\
\cline{2-12}
& 300 & 300 & 300 & 1542 & DDPM  & 61.46$\pm$2.33 & 43.95$\pm$5.43 & 83.10$\pm$1.98 & 78.22$\pm$3.41 & 46.10$\pm$7.91 & 42.87$\pm$7.45 \\
\cline{2-12}
& 350 & 350 & 350 & 1692 & DDPM  & \textbf{65.40$\pm$3.53} & 50.63$\pm$6.49 & 84.02$\pm$1.73 & 79.35$\pm$2.31 & 54.51$\pm$7.36 & 49.52$\pm$6.81 \\
\cline{2-12}
& 200 & 366 & 508 & 1716 & DDPM  & 62.63$\pm$1.17 & 45.38$\pm$5.84 & 82.47$\pm$1.19 & 77.56$\pm$1.99 & 45.39$\pm$7.49 & 43.87$\pm$6.79 \\
\cline{2-12}
& 400 & 400 & 400 & 2242 & DDPM  & 62.48$\pm$3.01 & 42.33$\pm$4.88 & 83.27$\pm$1.85 & 78.50$\pm$2.48 & 53.18$\pm$16.64 & 41.48$\pm$6.52 \\
\cline{2-12}
& 0   & 200 & 200 & 1042 & Blank & 63.50$\pm$1.66 & 40.68$\pm$1.53 & 83.67$\pm$0.57 & 79.18$\pm$1.12 & 37.89$\pm$1.68 & 38.12$\pm$1.75 \\
\cline{2-12}
& 0   & 250 & 250 & 1142 & Blank & 63.50$\pm$0.46 & 42.17$\pm$2.30 & 83.44$\pm$1.26 & 78.76$\pm$1.72 & 40.51$\pm$3.77 & 40.01$\pm$2.88 \\
\cline{2-12}
& 250 & 250 & 250 & 1392 & Blank & 63.21$\pm$2.04 & 41.67$\pm$3.41 & 83.60$\pm$0.43 & 78.87$\pm$1.09 & 38.17$\pm$2.77 & 38.99$\pm$3.78 \\
\cline{2-12}
& 350 & 350 & 350 & 1692 & Blank & 60.88$\pm$2.19 & 39.32$\pm$3.55 & 83.29$\pm$0.81 & 78.62$\pm$0.97 & 36.86$\pm$2.33 & 36.35$\pm$4.60 \\
\cline{2-12}
& 0   & 200 & 200 & 1042 & AvgDX & 60.73$\pm$1.17 & 39.47$\pm$3.15 & 82.03$\pm$1.17 & 76.55$\pm$2.40 & 38.54$\pm$4.16 & 36.55$\pm$3.73 \\
\cline{2-12}
& 0   & 250 & 250 & 1142 & AvgDX & 60.15$\pm$2.64 & 40.63$\pm$5.51 & 81.14$\pm$2.00 & 74.51$\pm$4.16 & 38.60$\pm$13.23 & 36.75$\pm$8.23 \\
\cline{2-12}
& 250 & 250 & 250 & 1392 & AvgDX & 62.63$\pm$3.40 & 42.71$\pm$3.96 & 83.69$\pm$1.60 & 78.96$\pm$1.56 & 44.44$\pm$5.00 & 41.40$\pm$5.34 \\
\cline{2-12}
& 350 & 350 & 350 & 1692 & AvgDX & 60.73$\pm$2.50 & 39.47$\pm$1.58 & 81.51$\pm$1.11 & 75.47$\pm$2.26 & 40.04$\pm$5.20 & 36.44$\pm$2.68 \\

\hline
\multirow{16}{*}{%
\parbox{3.2cm}{\centering
\textbf{T1 Only}\\
\textbf{(Train=642, Val=137, Test=137)}
}} 
& 0   & 0   & 0   & 642 & -- & 66.42$\pm$2.69 & 46.28$\pm$3.47 & 84.42$\pm$1.07 & 80.03$\pm$1.59 & 45.80$\pm$7.03 & 44.26$\pm$3.16 \\
\cline{2-12}
& 0   & 100 & 50  & 792 & -- & 66.57$\pm$2.82 & 49.24$\pm$7.13 & 85.87$\pm$1.08 & 82.18$\pm$1.33 & 60.39$\pm$9.93 & 48.63$\pm$7.48 \\
\cline{2-12}
& 0   & 100 & 100 & 842 & -- & 68.76$\pm$2.90 & 56.37$\pm$6.37 & 86.90$\pm$0.79 & 83.02$\pm$1.39 & 65.20$\pm$5.84 & 57.85$\pm$5.59 \\
\cline{2-12}
& 100 & 100 & 100 & 942 & -- & 66.57$\pm$1.87 & 51.08$\pm$5.35 & 86.16$\pm$0.89 & 82.91$\pm$0.85 & 65.76$\pm$6.01 & 51.07$\pm$5.41 \\
\cline{2-12}
& 0   & 200 & 100 & 942 & -- & 68.61$\pm$3.52 & 53.10$\pm$5.77 & 86.55$\pm$0.81 & 82.53$\pm$0.57 & \textbf{67.91$\pm$8.11} & 54.24$\pm$6.83 \\
\cline{2-12}
& 0   & 200 & 200 & 1042 & -- & 68.61$\pm$4.69 & 56.99$\pm$9.75 & 85.48$\pm$2.28 & 80.39$\pm$3.31 & 64.76$\pm$8.55 & 56.16$\pm$11.55 \\
\cline{2-12}
& 0   & 166 & 308 & 1116 & -- & 67.15$\pm$4.62 & 55.64$\pm$5.91 & 86.59$\pm$1.08 & 83.25$\pm$1.22 & 61.61$\pm$6.59 & 53.54$\pm$7.96 \\
\cline{2-12}
& 0   & 250 & 250 & 1142 & -- & 70.51$\pm$2.98 & 62.57$\pm$4.74 & 87.42$\pm$1.14 & 83.23$\pm$1.86 & 66.24$\pm$5.31 & 62.46$\pm$4.92 \\
\cline{2-12}
& 200 & 200 & 200 & 1242 & -- & 67.45$\pm$3.38 & 56.38$\pm$4.52 & 87.15$\pm$0.54 & 83.38$\pm$1.10 & 64.07$\pm$6.76 & 56.96$\pm$5.49 \\
\cline{2-12}
& 250 & 250 & 250 & 1392 & -- & 69.78$\pm$2.43 & 60.50$\pm$3.85 & 87.43$\pm$0.84 & 83.72$\pm$1.14 & 67.17$\pm$2.21 & 60.24$\pm$4.55 \\
\cline{2-12}
& 100 & 266 & 408 & 1416 & -- & 69.05$\pm$2.04 & 60.49$\pm$3.32 & 87.06$\pm$1.06 & 82.72$\pm$1.63 & 66.89$\pm$7.48 & 60.89$\pm$3.48 \\
\cline{2-12}
& 300 & 300 & 300 & 1542 & -- & 67.15$\pm$1.13 & 58.94$\pm$4.61 & 86.14$\pm$1.00 & 82.58$\pm$1.82 & 65.05$\pm$5.54 & 55.90$\pm$3.00 \\
\cline{2-12}
& 350 & 350 & 350 & 1692 & -- & 65.26$\pm$2.55 & 54.67$\pm$2.39 & 86.67$\pm$0.83 & 83.62$\pm$0.71 & 56.46$\pm$11.03 & 51.62$\pm$5.61 \\
\cline{2-12}
& 200 & 366 & 508 & 1716 & -- & 66.28$\pm$1.42 & 54.85$\pm$3.91 & 87.08$\pm$0.50 & \textbf{84.02$\pm$0.34} & 63.10$\pm$3.85 & 54.70$\pm$3.63 \\
\cline{2-12}
& 400 & 400 & 400 & 2242 & -- & \textbf{71.39$\pm$1.93} & \textbf{63.84$\pm$3.77} & \textbf{87.49$\pm$0.90} & 83.42$\pm$0.85 & 66.63$\pm$3.49 & \textbf{63.44$\pm$3.46} \\

\hline

\textbf{T1 Only*} & 0 & 0 & 0 & 3901 & -- & 68.91$\pm$4.20 & 62.70$\pm$1.23 & 84.56$\pm$3.17 & \textbf{84.28$\pm$0.84} & \textbf{71.60$\pm$6.05} & \textbf{64.14$\pm$2.59} \\
\cline{2-12}
\textbf{T1 Only$\dagger$} & 0 & 0 & 0 & 3901 & -- & 58.60$\pm$2.90 & 55.21$\pm$2.79 & 78.84$\pm$1.90 & 76.83$\pm$1.97 & 62.05$\pm$3.67 & 56.60$\pm$2.88 \\
\hline
\multicolumn{11}{l}{\footnotesize
Results are reported as mean $\pm$ standard deviation over 5 runs.
(* Train=3901, Test=137; $\dagger$ Train=3901, Test=859)}
\end{tabular}
}
\label{tab:results}
\end{table*}

\subsection{DWI-Only Model}
Augmenting the DWI-only training set with equal numbers of synthetic MCI and AD scans led to 
improvements across the evaluation metrics. The accuracy increased from $62.19\pm1.63$ to $65.40\pm2.19$, while balanced accuracy rose from $42.28\pm3.61$ to $51.68\pm7.06$. The Macro F1-score, which reflects the balance between precision and recall for each diagnosis, also showed a substantial increase from $41.00\pm5.20$ to $50.79\pm5.89$, suggesting improved detection of underrepresented diagnostic groups. Precision also improved notably, rising from $46.28 \pm 6.68$ to $57.18 \pm 9.07$ In contrast, stratified imputation, matching the original CN/MCI/AD class proportions using synthetic data did not produce notable improvements. Baseline imputation strategies such as blank and average imputation failed to improve performance and occasionally degraded it, suggesting that naive augmentation may introduce harmful biases.

\subsection{T1-Only Model}
%
No imputation was applied to the T1 modality due to the availability of abundant real T1 data. Instead, we examined the effect of increasing the volume of real T1 scans in a manner structurally analogous to the DWI imputation experiments. Training on the full T1 dataset ($n=3901$) and testing on the same set ($n=137$) used across experiments yielded high accuracy and AUC values. Further adjusting the class distribution to mimic imputation scenarios continued to improve performance, indicating that increased training diversity contributed positively. However, when the same trained model was evaluated on a much larger test set ($n=859$), performance decreased, likely due to increased heterogeneity and classification difficulty in the expanded evaluation cohort.

\subsection{Bimodal (T1 + DWI) Model}

In the bimodal T1+DWI model, augmenting the training set with synthetic MCI and AD scans increased accuracy from $68.03\pm2.33$ to $70.36\pm2.38$, and micro AUC increase from $84.99\pm1.43$ to $87.04\pm1.92$. On imputing equal amounts of CN, MCI and AD showed an increase in balanced accuracy from $46.13\pm2.26$ to $59.96\pm4.41$, and F1-score increase from $43.92\pm2.20$ to $58.67\pm7.55$. However, comparable improvements were achieved through blank imputation, suggesting that the performance gains arose primarily from increased exposure to real T1 scans rather than the synthetic DWI images. In contrast, average-diagnosis imputation produced no consistent improvements. These results imply that although DDPM-generated DWI contributed positively, its effect was overshadowed by the higher diagnostic value and larger volume of T1 data. This observation aligns with prior work showing that T1-weighted MRI often carries stronger discriminative information for AD and MCI classification than diffusion-derived measures \cite{agostinho2022combined}.

\section{Discussion}
We have performed extensive experiments to evaluate whether using conditional diffusion models (DDPM) for imputation of DWI from T1 scans can improve the accuracy of uni-modal and bi-modal deep learning models for
Alzheimer’s disease 3-way classification (cognitively normal, mild cognitive impairment, and Alzheimer’s
disease). Although we observe particular configurations with improvements in several metrics, particularly those sensitive to minority classes, there are no clear general patterns.

We plan to further explore these imputations techniques along three directions. First, we will train the diffusion-based image translator with a larger set of paired T1-DWI, which has recently become available. Second, we plan to explore newer diffusion bridge models \cite{zhang2025diffusion}. Finally, we will evaluate the methods both in simpler problems, such as binary AD classification, and harder problems, such as Parkinson's disease classification or cognitive score prediction, to understand whether the difficulty of the downstream evaluation task affects the usefulness.

\section{Acknowledgments}
This work was supported in part by NIH grants RF1AG081571, U01AG068057, R01MH131806, and S10OD032285. Special thanks to Wenbin Teng for help with the MONAI framework.

\bibliographystyle{IEEEtran}
\bibliography{refs}

\end{document}